# Design and development of an electronics-free earthworm robot

Riddhi Das[1,2][0009-0006-8320-8983], Joscha Teichmann[1,2][0000-0003-3501-3074], Thomas Speck[1,2][0000-0002-2245-2636] and Falk J. Tauber[1,2][0000-0001-7225-1472]

[1] Plant Biomechanics Group @ Botanical Garden, University of Freiburg, Germany
[2] Cluster of Excellence livMatS @ FIT–Freiburg Center for Interactive Materials and Bioinspired Technologies, University of Freiburg, Freiburg, Germany
`riddhi.das@livmats.uni-freiburg.de`

**Abstract.** Soft robotic systems have gained widespread attention due to their inherent flexibility, adaptability, and safety, making them well-suited for varied applications. Among bioinspired designs, earthworm locomotion has been extensively studied for its efficient peristaltic motion, enabling movement in confined and unstructured environments. Existing earthworm-inspired robots primarily utilize pneumatic actuation due to its high force-to-weight ratio and ease of implementation. However, these systems often rely on bulky, power-intensive electronic control units, limiting their practicality. In this work, we present an electronics-free, earthworm-inspired pneumatic robot utilizing a modified Pneumatic Logic Gate (PLG) design. By integrating preconfigured PLG units with bellow actuators, we achieved a plug-and-play style modular system capable of peristaltic locomotion without external electronic components. The proposed design reduces system complexity while maintaining efficient actuation. We characterize the bellow actuators under different operating conditions and evaluate the robot's locomotion performance. Our findings demonstrate that the modified PLG-based control system effectively generates peristaltic wave propagation, achieving autonomous motion with minimal deviation. This study serves as a proof of concept for the development of electronics-free, peristaltic soft robots. The proposed system has potential for applications in hazardous environments, where untethered, adaptable locomotion is critical. Future work will focus on further optimizing the robot design and exploring untethered operation using onboard compressed air sources.

## 1 Introduction

Soft robotic systems have gained significant attention in recent years due to their wide range of applications in search and rescue [1], medical robotics [2], human-robot interaction [3], studying bioinspired locomotion [4] and many more. The key advantage of soft robots lies in their material properties [5], which offer flexibility and adaptability, making them inherently safer than traditional rigid robots while also reducing computational complexity [6]. While rigid robots have demonstrated remarkable capabilities in various environments, they face challenges in navigating highly constricted unpredicted environments. Nature provides numerous examples of organisms that have evolved to survive and inhabit in such extreme conditions, inspiring researchers to develop bioinspired robotic systems [7–9]. Among these, earthworm locomotion has been extensively studied due to its efficient peristaltic motion, which allows movement in confined and unstructured environments.

Earthworms are entirely soft-bodied invertebrates that utilize a hydrostatic skeleton [10, 11] consisting of fluid-filled [12] segments enclosed by two antagonistic muscle groups: longitudinal and circular muscles. The sequential actuation of these muscle groups enables each segment to either elongate longitudinally or expand radially. The propagation of these deformations along the body generates a peristaltic wave [13], which serves as the primary mechanism for locomotion. In addition, earthworms use bristle-like structures called setae [14] on their ventral side to generate anisotropic friction, further aiding movement.

Several earthworm-inspired robotic systems have been developed with applications in endoscopy [15], pipe inspection [16], and planetary exploration [17]. Pneumatic actuation [18] has been a popular choice in these designs due to its high force-to-weight ratio and ease of implementation. Notable examples include the work of Liu et al. [19], where kirigami scales were integrated into a pneumatic robot to enhance friction, and of Li et al. [20], who developed a steerable earthworm-inspired robot actuated by both positive pressure and vacuum. Similarly, Das et al. [21, 22] created an earthworm-inspired robot that demonstrated locomotion across various terrains.

Despite their advantages, pneumatically actuated robots often rely on bulky, power-intensive control systems, limiting their practicality. A major breakthrough in addressing this challenge has been the development of soft valves [23]. While early fabrication techniques for soft valves were complex and labor-intensive, advancements in additive manufacturing have enabled their effortless integration into robotic designs [24]. These innovations have led to the emergence of electronics-free soft robots[25, 26] equipped with onboard control and sensing systems. Such robots typically require only a single compressed air source for actuation and can be further adapted for untethered operation using compressed air canisters. However most of the state-of-the-art electronics free robots suffer from complexity and do not allow multiple modules to be compactly integrated.

Several researchers have explored the potential of soft valve-based robotic systems. Conrad et al. [24] developed Pneumatic Logic Gates (PLG) and integrated them into a monolithic compliant walker. Teichmann et al. [27] controlled an insect-inspired soft robot with a pneumatic circuit using PLG to make it walk according to the locomotion rules of stick insects. Similarly, Drotman et al. [28] designed an electronics-free pneumatic walker using bellow actuators as legs. Jiao et al. [29] introduced a modular LEGO-like soft robot incorporating Kresling units capable of sensing stimuli, storing data, and performing locomotion.

In this work, we present a novel electronics-free, earthworm-inspired pneumatic robot based on a modified PLG design. Building upon the PLG units developed by Conrad et al. [24], we introduce design modifications that preconfigure the operation, thereby reducing complexity, and enabling seamless series interconnection. These modified PLG units are integrated with a bellow actuator to create a fully 3D-printable robot module. The novelty of this work lies in the valve structure and the development of a module with an integrated PLG design, which enables easy modular assembly, resulting in a soft robotic system capable of peristaltic locomotion without the need for electronic components. This study serves as a proof of concept for fully integrated, peristaltic soft robotic systems, with potential applications in search and rescue operations in hazardous environments.

## 2    Materials and Methods

### 2.1    Modified PLG Design

The Pneumatic Logic Gate (PLG) units, as introduced in Conrad et al.[24], function analogously to electronic MOSFETs. As shown in Fig. 1a, the PLG structure integrates two valves, $V_1$ and $V_2$, which operate as a normally open (NO) and a normally closed (NC) valve, respectively. This configuration enables the PLG to function as a pressure-activated three/two-way valve. Depending on the input pressure conditions (HIGH/LOW), the PLG unit can perform various Boolean operations, including NOT, OR, and AND. In this context, HIGH refers to a state analogous to boolean 1, where the supplied air pressure exceeds the minimum operating pressure of the PLG. Conversely, LOW corresponds to boolean 0, where the inlets remain at atmospheric pressure or a pressure lower than the minimum operating threshold. The Boolean logic demonstrations were conducted using two of the three sockets $S_{C1}$, $S_{C2}$ and $S_T$ as input depending on the desired operation and $S_{OUT}$ as the output.

As depicted in Fig. 1a, the $S_{p+}$ inlet supplies a constant positive pressure (HIGH signal) to maintain $V_2$ in its NC state. The $S_T$ inlet serves as the trigger signal, while $S_{OUT}$ represents the output of the PLG. Additionally, the inlets $S_{C1}$ and $S_{C2}$ regulate airflow through V1 and V2, respectively, directly influencing $S_{OUT}$, as demonstrated in Conrad et al. [24]. In this study, the PLG design was modified to two distinct units: an inverter, shown in Fig. 1b, and a buffer, shown in Fig. 1c. To facilitate series interconnectivity, the inlets and outlets of the PLG were reconfigured. As illustrated in Fig. 1b and 1c, the $S_{p+}$ inlet was modified into a T-connection on the extreme left to actuate $V_2$ and subsequent modules. Similarly, for streamlined connectivity, the trigger inlet ($S_T$) was relocated to the extreme right. The output ($S_{OUT}$) was split into two channels—one directed to actuate the bellow actuator (discussed in detail in the following section) and the other aligned with the $S_T$ axis.

As the name suggests, the inverter module inverses the input signal. If the PLG is triggered with a HIGH input signal ($S_T$), the output ($S_{OUT}$) will be LOW, and vice versa. In the previous PLG design, in addition to the constant $S_{p+}$ supply, a HIGH signal at $S_{C1}$ was required to function as an inverter. In the modified PLG inverter design, the $S_{p+}$ and $S_{C1}$ were merged to simultaneously pressurize the tube in $V_1$ and the side chambers of $V_2$, as depicted in Fig. 1b. This modification simplifies the inverter's operation. The operation of the inverter module and its corresponding truth table is presented in

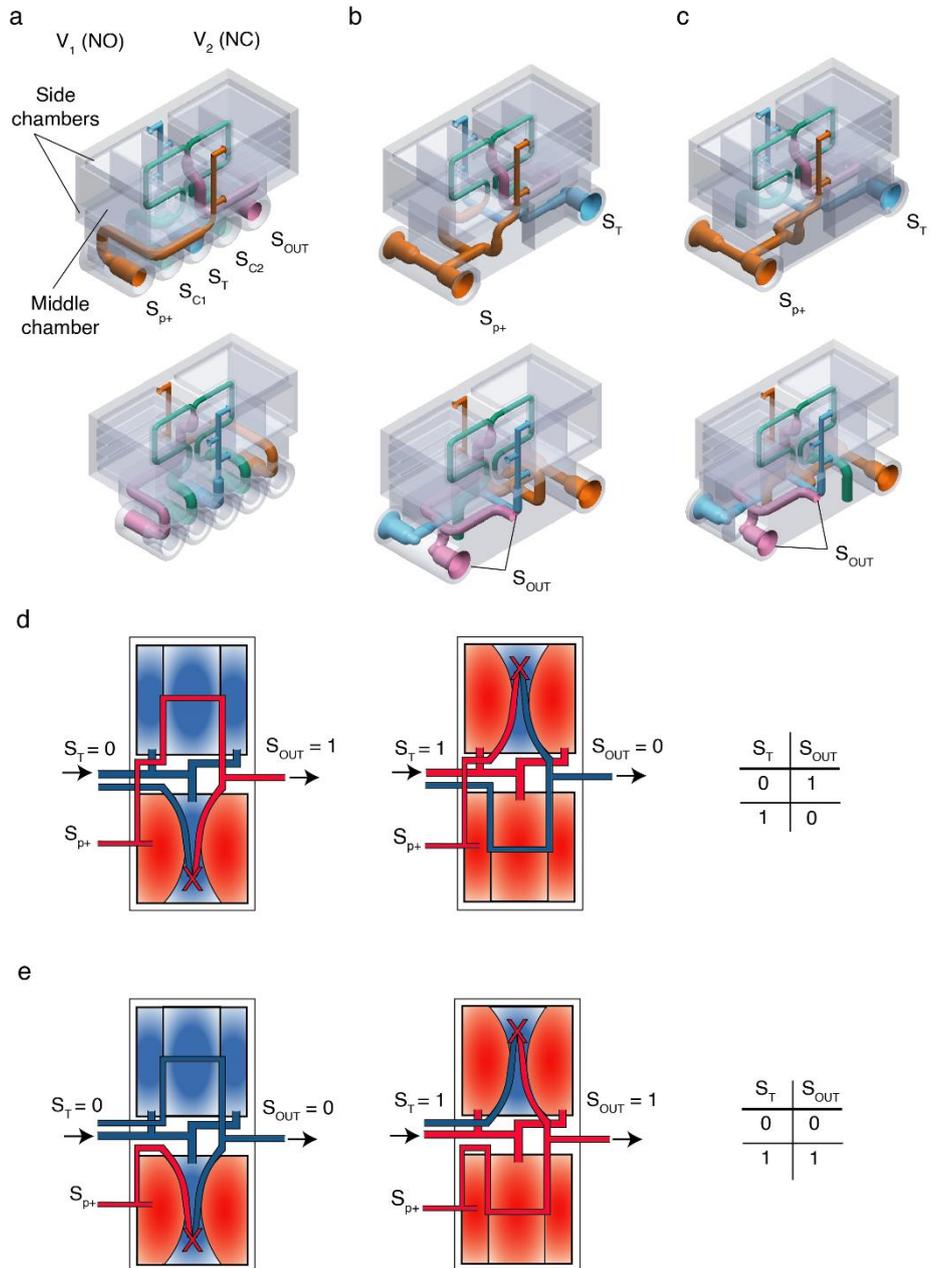

**Fig. 1.** Modified PLG Design: (a) Isometric front and back views of the conventional PLG. (b) Isometric front and back views of the modified PLG inverter. (c) Isometric front and back views of the modified PLG buffer. (d) Operational diagram and truth table of the modified PLG inverter, demonstrating signal inversion at the output. (e) Operational diagram and truth table of the modified PLG buffer, illustrating the passage of the trigger signal to the output.

Fig. 1d. The buffer module transmits the same signal, but disconnects the pneumatic circuit between input and output, refreshes the signal and introduces a delay. To pre-configure the buffer module, the $S_{p+}$ and $S_{C2}$ were merged to pressurize both the tube in $V_2$ and the side chambers of $V_2$, as shown in Fig. 1c. The working principle of the modified buffer module and its corresponding truth table are illustrated in Fig. 1e. In both modules, the socket for the evacuation of pressure has been omitted and the exhaust channel terminates at the module wall.

## 2.2   Robot Module Design

Linear elongation and contraction are essential for generating a peristaltic wave in soft robotic systems. Bellow-based actuators are a common solution in pneumatic soft robots to achieve this motion. As shown in Fig. 2a, the designed bellow features a pitch of 2.65 mm, an external angle of 50°, an internal angle of 61°, and a wall thickness at the valleys of 1.6 mm. A decrease in the bellow pitch can lead to higher deformation. However, further reduction in the internal and external bellow angles were observed to cause sagging. This sagging ultimately resulted in print failure due to the challenges associated with printing steep overhangs. The external diameter and the length of the bellow are 28 mm and 33 mm, respectively.

To develop a peristaltic module, the bellow design was integrated with a modified PLG unit, as shown in Fig. 2b. The entire module, including the bellow actuator, middle cap, end cap, and modified PLG unit, was 3D printed monolithically using Thermoplastic Polyurethane with Shore A 70 hardness (TPU A70) on a Prusa i3 Mk3s printer (Prusa, Prague, Czech Republic) modified as detailed by Conrad et al. [30] with an average print time of 16 hours. The caps were designed with small slots on diagonally opposite sides to enable connection between successive modules. Additionally, apertures were incorporated in the caps to allow the passage of pneumatic tubes.

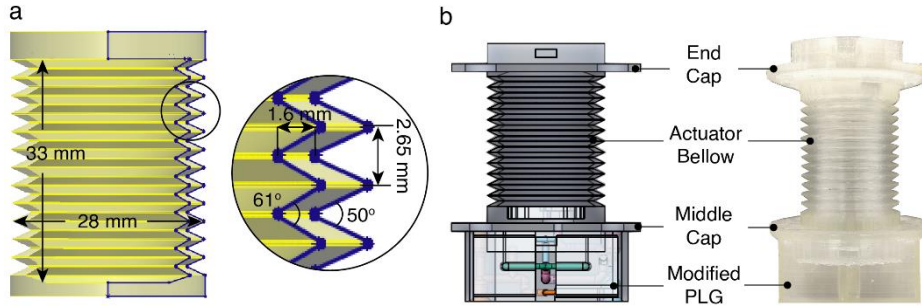

**Fig. 2.** Robot Module Design: (a) Internal dimensions of the bellow actuator. (b) Fully 3D-printed robot module with its individual components: Modified PLG, Middle Cap, Bellow Actuator, and End Cap

## 2.3   Robot Assembly

The goal of the robot assembly was to generate a continuous wave pattern to propel the robot forward. As shown in Fig. 3a, the pneumatic circuit was developed using four modules. The initiator inverter module consists of a modified PLG functioning as an inverter, followed by four buffer modules consisting of modified PLG functioning as buffers. On the left side along the longitudinal axis of the robot, a positive pressure supply ($S_{p+}$) was provided to the inverter module, forming a continuous supply line for the subsequent buffer modules. A pneumatic stopper was placed at the fourth module to block airflow. On the right side of the longitudinal axis, the output of the initial inverter module was connected to the trigger ($S_T$) of the next buffer module, and this connection pattern was maintained across the subsequent modules. The output of the fourth module was looped back to the input of the inverter module, forming a pneumatic ring oscillator circuit that enabled continuous operation.

With a constant pressure supply, the modified PLG in all modules were activated. Initially, the inverter module outputs a HIGH signal, since its $S_T$ input is LOW. This HIGH signal at $S_{OUT}$ inflates the bellow integrated within the inverter module and simultaneously serves as the $S_T$ input for the next buffer module. The buffer module, upon receiving a HIGH $S_T$ input, also generates a HIGH $S_{OUT}$ output, inflating its bellow. This process continues through the subsequent two buffer modules, with each HIGH $S_T$ input leading to a HIGH $S_{OUT}$ output and bellow actuation. When the fourth and final module outputs a HIGH signal, it is fed back as the $S_T$ input to the inverter module. Upon receiving this HIGH signal, the inverter module switches its output to LOW, deflating its bellow and inverting the states of all subsequent modules. As a result, all actuated bellows release pressure, completing the cycle. This alternating pattern of inflation and deflation continuously generates peristaltic waves, as shown in Fig. 3b.

The robot assembly was achieved by joining individual modules using clamp-like interconnections, as illustrated in Fig. 3c. Small slots in the module end caps facilitated the insertion of these interconnections, ensuring a secure fit. The bottom portion of the interconnections featured setae-like passive structures, providing anisotropic friction to aid locomotion. The pneumatic circuit was interconnected using pneumatic connectors (3 mm external diameter) and Polyurethane (PU) tubes of length 140 mm, which linked the inlets and outlets of each module. The tube lengths were deliberately kept longer than the spacing between modules to accommodate the expansion of the robot during actuation. The 3D model and prototype of the fully assembled robot (length: 263 mm, weight: 191 g), along with its pneumatic connections, are depicted in Fig. 3d and 3e, respectively.

### 2.4 Experimental Methodology

To develop the robot, individual bellow actuators were initially characterized to evaluate their performance and determine the optimal governing parameters. Based on the characterization results, the final design of the individual modules was established and fabricated using 3D printing. Following the manufacturing process, the assembled robot was tested to assess its locomotion performance.

**Actuator Characterization**

To evaluate the maximum elongation of an individual actuator module, a series of characterization experiments were conducted. Five bellow actuators were fabricated using 3D printing, with three different wall thicknesses (1.3 mm, 1.6 mm, and 1.9 mm). A robust experimental setup, as shown in Fig. 4a, was designed to securely clamp the actuators and facilitate deformation measurements. A red marker was placed on the top of each bellow actuator, and its movement was recorded using a high-speed camera.
The actuators were supplied with constant pressures of 1.7 bar, 2.0 bar, and 2.3 bar, chosen based on the optimal operating pressure of the PLGs. Cyclic deformation experiments were performed at different actuation durations (AD) with intervals of 200 ms, ranging from 200 ms to 1000 ms, while the release time was maintained at 800 ms across all experiments. The actuation parameters were precisely controlled using a custom pneumatic control system [31]. The recorded deformation videos were analyzed using Tracker (V 6.2.0, Open Source Physics, Indianapolis, USA), an open-source motion-tracking software, and the extracted raw data was processed and visualized using Python.

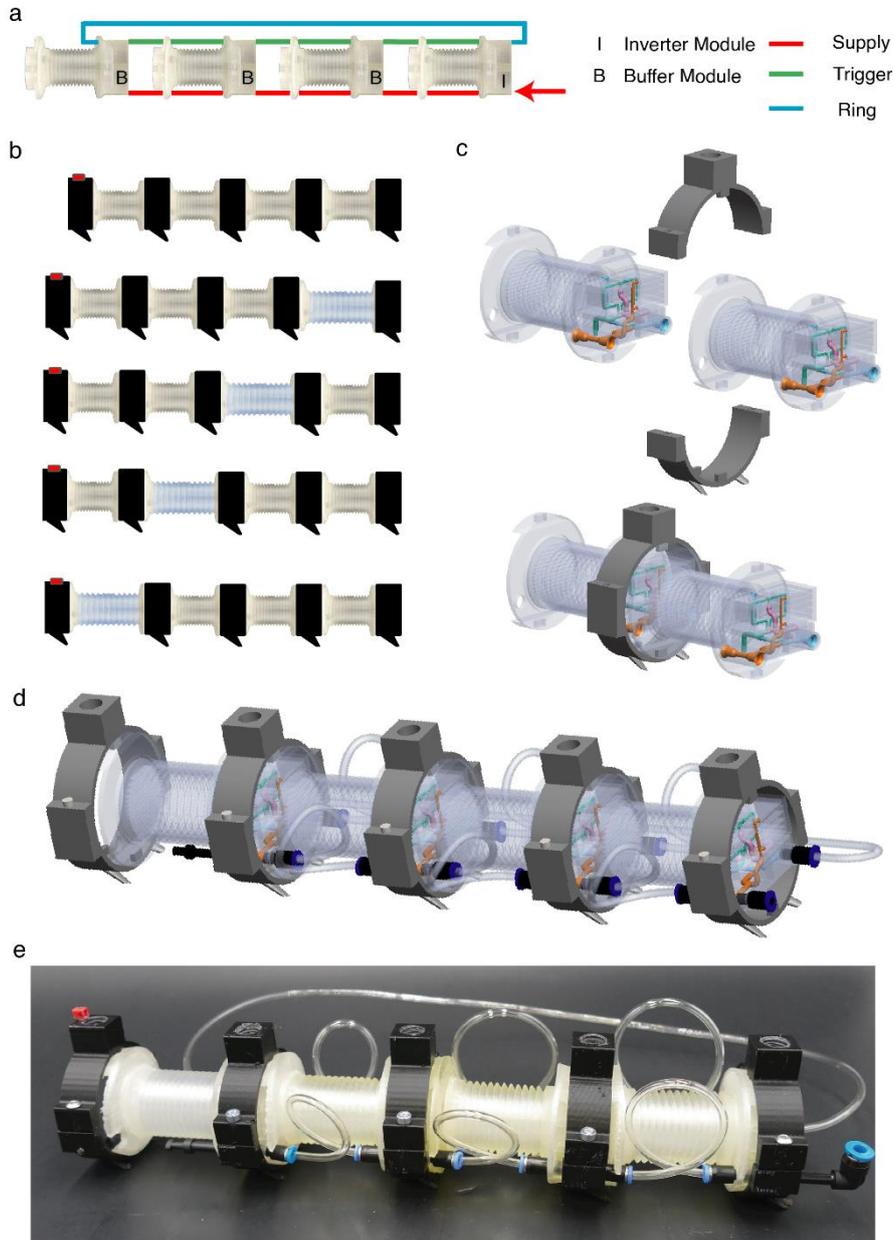

**Fig. 3.** Peristaltic Robot Assembly: (a) Pneumatic circuit diagram illustrating the ring oscillator formed by robot modules to generate a continuous peristaltic wave. (b) Schematic representation of peristaltic wave propagation through the robot during locomotion. (c) Attachment mechanism of individual robot modules using clamp-like interconnections. (d) CAD model of the fully assembled robot. (e) Physical prototype of the fully assembled robot.

**Robot locomotion**

The locomotion performance of the assembled robot was evaluated on a polyethylene foam surface under a constant pressure of 2 bar. A marker was placed on the tip of the robot, and its motion was recorded using both overhead and front-facing cameras. An additional experiment was conducted to further analyze the inverter and buffer modules switching frequency during robot locomotion. Thus, the outputs of each module before connecting to the trigger of the subsequent module as shown in the pneumatic circuit, was split and connected to separate pressure sensors. Data from these sensors were continuously recorded throughout the entire locomotion period, providing insights into the pressure variations and actuation dynamics of the system.

## 3 Results and Discussion

### 3.1 Actuator Characterization

Accurately estimating the elongation of the bellow actuator is crucial in designing an optimal robot module. A comparative analysis of actuator displacement versus AD, as illustrated in Fig. 4b, 4c, and 4d, reveals that actuators with a bellow thickness of 1.9 mm exhibit the least deformation.
At 2.0 bar, actuators with a 1.6 mm wall thickness demonstrated average deformations of 9.2 mm, 10.8 mm, 11.6 mm, and 11.9 mm for AD of 200 ms, 400 ms, 600 ms, and 800 ms, respectively. At an increased pressure of 2.3 bar, the same actuators exhibited deformations of 10.1 mm, 12.3 mm, 13.0 mm, and 14.0 mm for the respective AD. The displacement trends observed in Fig. 4c and 4d indicate a gradual increase in deformation when the AD increases from 200 ms to 400 ms, after which the deformation curve plateaus. This suggests that the optimal AD is approximately 400 ms.
Actuators with a 1.3 mm wall thickness exhibited a rapid increase in deformation with increasing AD. However, at pressures of 2.0 bar and 2.3 bar, they experienced material fatigue and deformation failure when actuated at 1000 ms, indicating a loss of elastic recovery. Although these actuators demonstrated higher initial deformations, their failure at prolonged AD suggests a material creep effect, leading to nonlinear deformation over time. Consequently, the bellow thickness must be greater than 1.3 mm to ensure long-term durability. Conversely, actuators with a 1.9 mm thickness displayed significantly lower deformations compared to the 1.6 mm variant, making them less effective for achieving sufficient displacement. Based on these observations, a 1.6 mm bellow thickness was determined to be the most optimal choice for the robot module, in terms of material elasticity, thickness and operation pressure.

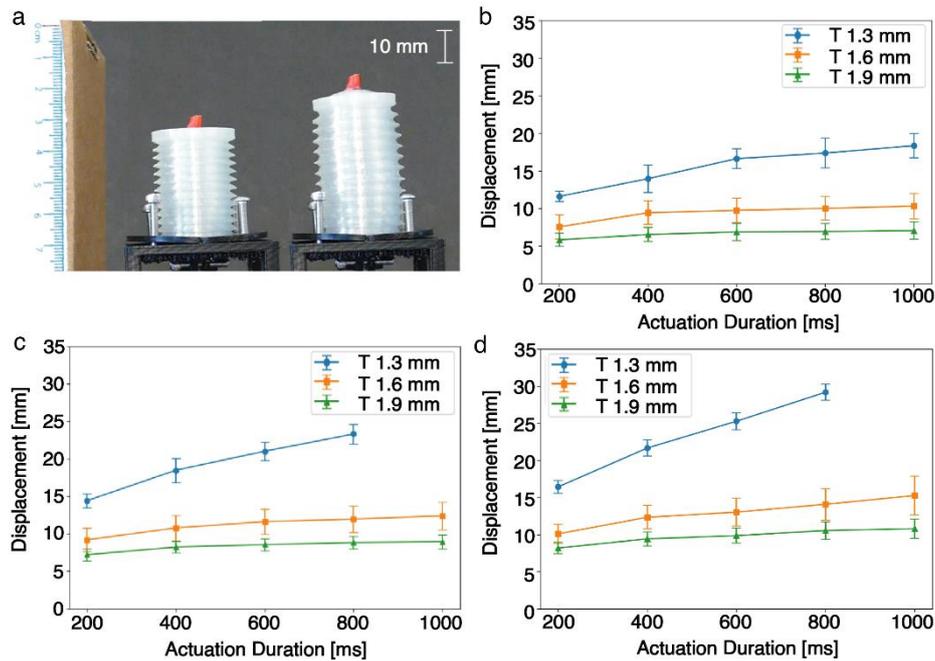

**Fig. 4.** Actuator Characterization Results: (a) Experimental setup for linear displacement characterization of the bellow actuator. (b) Linear displacement vs. actuation duration (AD) at 1.7 bar, demonstrating that actuators with greater bellow thickness (T) exhibit lower deformation. (c) Linear displacement vs. AD at 2.0 bar, showing a rapid increase in deformation for T = 1.3 mm, which ultimately fails at AD = 1000 ms. (d) Linear displacement vs. AD at 2.3 bar, confirming similar failure behavior for T = 1.3 mm, indicating its unsuitability for use as a module at higher pressures.

### 3.2 Robot Locomotion

As shown in Fig. 5a and 5b, the robot generated a peristaltic wave pattern, achieving a locomotion velocity of 4.03 mm/s, which approximates to 0.015 body lengths per second. Over the course of its locomotion, the robot exhibited a cumulative deviation of 26 mm along the axis perpendicular to its direction of motion. This minimal deviation can be attributed to the nonlinearity of the actuator material or to the stiff pneumatic supply line.

The modified PLG state switching time for the entire wave pattern was calculated to be 5.98 s (0.167 Hz). As observed in Fig. 5c, although the maximum pressure recorded at the outlet of the modules M1, M2 and M3 was around 1.55 bar, in module M4 a maximum pressure of 1.85 bar was recorded. The higher activation pressure observed in this buffer module could be due to minor leakages within the assembled system, which are challenging to completely eliminate.

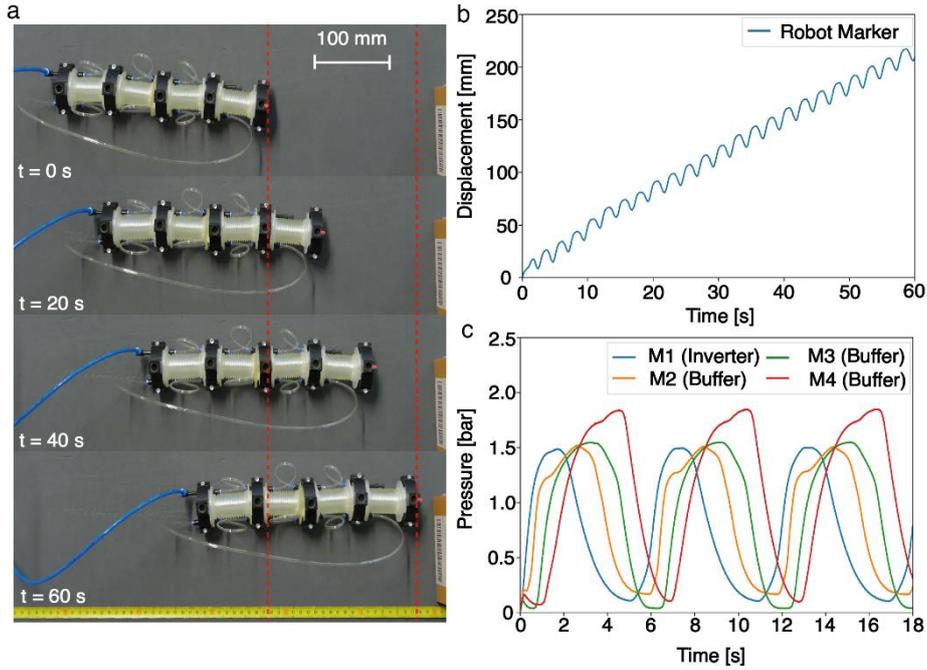

**Fig. 5.** Robot Locomotion Results: (a) Sequential timestamps capturing the robot's movement over time (b) Displacement of the robot's marker over time, indicating a locomotion velocity of 4.03 mm/s (c) Pressure vs. time plot of the modified PLG state switching, showing a maximum pressure of 1.55 bar for the first three modules, M1 (Inverter), M2 (Buffer), M3 (Buffer) and 1.85 bar for the fourth module M4(Buffer).

## 4   Conclusion

This study presents an electronics-free, earthworm-inspired pneumatic robot that utilizes a modified Pneumatic Logic Gate (PLG) system for peristaltic locomotion. By integrating preconfigured PLG units with bellow actuators, the proposed design enables modular assembly while eliminating the need for electronic components. Actuator characterization experiments demonstrated that a bellow thickness of 1.6 mm provides optimal deformation for effective motion. The locomotion analysis confirmed the successful generation of a peristaltic wave, achieving a velocity of 4.03 mm/s with minimal deviation. The use of soft valves and pneumatic actuation reduces system complexity while maintaining efficient and adaptive locomotion, making this approach promising for future applications in search and rescue operations in confined environments. Compared to conventional pneumatically actuated robots, the elimination of bulky electronic control units enhances practicality and portability. The modular design of the peristaltic module enables seamless interconnection of successive modules and allows for easy manual switching of pneumatic connections to change the operational wave pattern. Future work will focus on optimizing the robot design for the study of different wave patterns, incorporating retractable setae, and developing an obstacle-avoidance, direction-changing tip module. Additionally, transitioning to an untethered system using onboard compressed air canisters could further enhance the robot's applicability in real-world scenarios. This research serves as a proof of concept for fully integrated, peristaltic soft robotic systems, contributing to the advancement of bio-inspired soft robotics.

**Acknowledgments.** Funded by the Deutsche Forschungsgemeinschaft (DFG, German Research Foundation) under Germany's Excellence Strategy – EXC-2193/1 – 390951807.

**Disclosure of Interests.** The authors have no competing interests to declare that are relevant to the content of this article.

# References

1. Hawkes, E.W., Blumenschein, L.H., Greer, J.D., Okamura, A.M.: A soft robot that navigates its environment through growth. Sci Robot. 2, 1–8 (2017).
2. Park, C., Ozturk, C., Roche, E.T.: Computational Design of a Soft Robotic Myocardium for Biomimetic Motion and Function. Adv Funct Mater. 32, (2022).
3. Casas-Bocanegra, D., Gomez-Vargas, D., Pinto-Bernal, M.J., Maldonado, J., Munera, M., Villa-Moreno, A., Stoelen, M.F., Belpaeme, T., Cifuentes, C.A.: An open-source social robot based on com-pliant soft robotics for therapy with children with ASD. Actuators. 9, (2020).
4. Das, R., Murali Babu, S.P., Palagi, S., Mazzolai, B.: Soft Robotic Locomotion by Peristaltic Waves in Granular Media. 2020 3rd IEEE International Conference on Soft Robotics, RoboSoft 2020. 223–228 (2020).
5. Chubb, K., Berry, D., Burke, T.: Towards an ontology for soft robots: what is soft? Bioinspir Biomim. 14, 063001 (2019).
6. Hauser, H., Hughes, J.: Morphological computation—Past, present and future, (2024).
7. Hosoi, A.E., Goldman, D.I.: Beneath our feet: Strategies for locomotion in granular media. Annu Rev Fluid Mech. 47, 431–453 (2015).
8. Calisti, M., Giorelli, M., Levy, G., Mazzolai, B., Hochner, B., Laschi, C., Dario, P.: An octopus-bioinspired solution to movement and manipulation for soft robots. Bioinspir Biomim. 6, (2011).
9. Zhao, D., Luo, H., Tu, Y., Meng, C., Lam, T.L.: Snail-inspired robotic swarms: a hybrid connector drives collective adaptation in unstructured out-door environments. Nat Commun. 15, (2024).
10. CHAPMAN, G.: the Hydrostatic Skeleton in the invertebrates. Biological Reviews. 33, 338–371 (1958).
11. Kier, W.M.: The diversity of hydrostatic skeletons. Journal of Experimental Biology. 215, 1247–1257 (2012).

12. Newell, B.Y.G.E.: The Role of the Coelomic Fluid in the Movements of Earthworms. Journal of Experimental Biology. 27, 110–122 (1950).
13. Gray, J., Lissmann, H.W.: Vii . Locomotory Reflex-es in the Earthworm. Journal of Experimental Biology. 15, 506–517 (1938).
14. Samal, S., Samal, R.R., Mishra, C.S.K., Sahoo, S.: Setal anomalies in the tropical earthworms *Drawida willsi* and *Lampito mauritii* exposed to elevated concentrations of certain agrochemicals: An electron micrographic and molecular docking approach. Environ Technol Innov. 15, (2019).
15. Mangan, E.V, Kingsleyl, D.A., Quinnl, R.D., Chie, H.J.: Development of a Peristaltic Endoscope. Building. 347–352 (2002).
16. Liu, X., Song, M., Fang, Y., Zhao, Y., Cao, C.: Worm-Inspired Soft Robots Enable Adaptable Pipeline and Tunnel Inspection. Advanced Intelligent Systems. 2100128, (2021).
17. Isaka, K., Tsumura, K., Watanabe, T., Toyama, W., Okui, M., Yoshida, H., Nakamura, T.: Soil Dis-charging Mechanism Utilizing Water Jetting to Im-prove Excavation Depth for Seabed Drilling Explorer. IEEE Access. 8, 28560–28570 (2020).
18. Boyraz, P., Runge, G., Raatz, A.: An overview of novel actuators for soft robotics. High Throughput. 7, 1–21 (2018).
19. Liu, B., Ozkan-Aydin, Y., Goldman, D.I., Hammond, F.L.: Kirigami skin improves soft earthworm robot anchoring and locomotion under cohesive soil. RoboSoft 2019 - 2019 IEEE International Conference on Soft Robotics. 828–833 (2019).
20. Li, P., Chen, B., Liu, J.: Multimodal steerable earthworm-inspired soft robot based on vacuum and positive pressure powered pneumatic actuators. Bioinspir Biomim. 19, (2024).
21. Das, R., Babu, S.P.M., Visentin, F., Palagi, S., Mazzolai, B.: An earthworm-like modular soft robot for locomotion in multi-terrain environments. Sci Rep. 13, (2023).
22. Das, R., Babu, S.P.M., Mondini, A., Mazzolai, B.M.: Effects of lateral undulation in granular medium burrowing with a peristaltic soft robot. In: 2023 IEEE International Conference on Soft Robotics, RoboSoft 2023. Institute of Electrical and Electronics Engineers Inc. (2023).
23. Rothemund, P., Ainla, A., Belding, L., Preston, D.J., Kurihara, S., Suo, Z., Whitesides, G.M.: A soft, bi-stable valve for autonomous control of soft actuators. Sci. Robot.3, eaar7986 (2018).
24. Conrad, S., Teichmann, J., Auth, P., Knorr, N., Ulrich, K., Bellin, D., Speck, T., Tauber, F.J.: 3D-printed digital pneumatic logic for the control of soft robotic actuators. Sci. Robot.9, eadh4060 (2024).
25. Miyaki, Y., Tsukagoshi, H.: Self-Excited Vibration Valve That Induces Traveling Waves in Pneumatic Soft Mobile Robots. IEEE Robot Autom Lett. 5, 4133–4139 (2020).
26. Xu, K., Perez-Arancibia, N.O.: Electronics-Free Logic Circuits for Localized Feedback Control of Multi-Actuator Soft Robots. IEEE Robot Autom Lett. 5, 3990–3997 (2020).
27. Teichmann, J., Auth, P., Conrad, S., Speck, T., Tauber, F.J.: An insect-inspired soft robot controlled by soft valves. In: Lecture Notes in Computer Science (including subseries Lecture Notes in Artificial Intelligence and Lecture Notes in Bioinformatics). pp. 428–441. Springer Science and Business Media Deutschland GmbH (2023).
28. Drotman, D., Jadhav, S., Sharp, D., Chan, C., Tolley, M.T.: Electronics-free pneumatic circuits for controlling soft-legged robots. Sci Robot. 6, (2021).
29. Jiao, Z., Hu, Z., Shi, Y., Xu, K., Lin, F., Zhu, P., Tang, W., Zhong, Y., Yang, H., Zou, J.: Reprogrammable, intelligent soft origami LEGO coupling ac-tuation, computation, and sensing. Innovation. 5, (2024).
30. Conrad, S., Speck, T., Tauber, F.J.: Tool changing 3D printer for rapid prototyping of advanced soft robotic elements. Bioinspir Biomim. 16, (2021).

31. Esser, F., Krüger, F., Masselter, T., Speck, T.: Characterization of Biomimetic Peristaltic Pumping System Based on Flexible Silicone Soft Robotic Actuators as an Alternative for Technical Pumps. In: Lecture Notes in Computer Science (including subseries Lecture Notes in Artificial Intelligence and Lecture Notes in Bioinformatics). pp. 101–113. Springer Verlag (2019).